\documentclass[letterpaper, 10 pt, conference]{ieeeconf}  

\IEEEoverridecommandlockouts                              

\usepackage{graphics} 
\usepackage{epsfig} 
\usepackage{amsmath} 
\usepackage{amssymb}  
\usepackage{hyperref}
\usepackage{url}
\usepackage{booktabs}
\usepackage{multirow}

\title{\LARGE \bf
FP-Loc: Lightweight and Drift-free \\ Floor Plan-assisted LiDAR Localization
}

\author{Ling Gao$^{1,2,3}$ and Laurent Kneip$^{1,4}$
\thanks{$^{1}$ShanghaiTech University. $^{2}$Shanghai Institute of Microsystem and Information Technology, Chinese Academy of Sciences. $^{3}$University of Chinese Academy of Sciences. $^{4}$Shanghai Engineering Research Center of Intelligent Vision and Imaging. \tt\url{https://mpl.sist.shanghaitech.edu.cn}}%
}

\begin{document}

\maketitle
\thispagestyle{empty}
\pagestyle{empty}


\begin{abstract}
We present a novel framework for floor plan-based, full six degree-of-freedom LiDAR localization. Our approach relies on robust ceiling and ground plane detection, which solves part of the pose and supports the segmentation of vertical structure elements such as walls and pillars. Our core contribution is a novel nearest neighbour data structure for an efficient look-up of nearest vertical structure elements from the floor plan. The registration is realized as a pair-wise regularized windowed pose graph optimization. Highly efficient, accurate and drift-free long-term localization is demonstrated on multiple scenes.
\end{abstract}


\section{Introduction}

Robust, accurate, and efficient localization is the back-bone of many mobile autonomous systems such as smart vehicles and robots. In this context, Simultaneous Localization And Mapping (SLAM) is often pro-claimed as an essential algorithm to be run on data captured incrementally by exteroceptive sensors such as LiDARs or cameras, especially in GPS-denied indoor environments. SLAM relies on the assumption that no prior knowledge about the environment is given, and a map of the surrounding scene needs to be constructed alongside tracking the pose.

This assumption often does not have to be made. Priors about the architecture of indoor environments may often be available and given for example in the form of a dense 3D point cloud. The construction of such point clouds is however expensive and requires accurate large-scale SLAM with loop closures and semantic annotations to be run upfront. High-end devices that can be used for this task are offered by FARO~\cite{faro}, GeoSLAM~\cite{geoslam}, or NavVis~\cite{navvis}.

A much cheaper prior for the geometry of indoor environments is given by building floor plans, which are often readily available from the construction phase or Building Information Models (BIM). We present the first approach to perform full six degree-of-freedom tracking of 3D LiDARs in indoor environments using floor plan priors. Our method relies on the existence of horizontal ceiling and ground planes, and robustly measures those planes in 3D LiDAR scans. Ceiling and ground floor plane sensing is used to identify the roll and pitch angles as well as the height of the sensor. We furthermore use this information to compensate for rotation and extract vertical wall features from the scans. The wall points are finally projected onto a horizontal plane, after which they can be aligned with the features of a floor plan. We assume that an initial pose is given and focus on continuous tracking of the absolute pose of the sensor.

Our contributions are listed as follows:

\begin{itemize}
    \item We propose a novel data structure in the form of an approximate nearest neighbour field (ANNF) which enables an efficient look-up of the nearest geometric floor plan elements (e.g. wall segments) for any given point in the localization space. This method greatly assists in reducing the required computation time for correspondence matching while maintaining high success rate in retrieving the nearest neighbour.
    \item We design a LiDAR indoor localization algorithm using our newly proposed ANNF structure, which achieves drift-free six degree-of-freedom tracking performance.
    \item We evaluate our system on benchmark datasets across various texture-less environments under different motion speeds. The comparison results not only show that our algorithm is suitable for long-term localization in challenging scenarios such as long corridors, but also outperforms traditional SLAM-based algorithms in computational efficiency. It runs at 5x real-time on a laptop, and thus remains applicable on embedded hardware.
\end{itemize}


\section{Related Work}

LiDAR-based localization and mapping is often solved using LOAM-based odometry algorithms~\cite{zhang2014loam, zhang2017low, shan2018lego, ye2019tightly, shan2020lio}. Instead of using fundamental geometric methods such as the point-to-point ICP algorithm \cite{besl1992method} or one of its variants (e.g. point-to-plane ICP \cite{rusinkiewicz2001efficient}, Generalized-ICP \cite{segal2009generalized}), the LOAM paradigm consists of downsampling the point clouds by extracting distinctive features. Feature-based methods are favored over dense ICP variants owing to their robustness and efficiency. Furthermore, the use of matched features enables the construction of optimization problems over poses and landmarks, similar to what bundle adjustment achieves for regular camera-based solutions. Note that many of the aforementioned frameworks perform loose coupling with an Inertial Measurement Unit (IMU) for motion compensation in LiDAR readings and motion priors for the registration of consecutive frames \cite{zhang2014loam, zhang2017low, shan2018lego}. Tight coupling with pre-integrated IMU readings and thus better accuracy is achieved in \cite{ye2019tightly, shan2020lio}. In addition, \cite{shan2020lio} also takes GPS signals into account in order to support drift-free outdoor localization and mapping.

For absolute indoor localization, the above methods are not an option as they are lacking absolute reference sensor readings in those environments. Several alternatives based on RFID, WiFi, Bluetooth, or ultra-sonic beacons exist. A complete review of the various technologies is given in \cite{obeidat2021review}.

In the present work we use prior map information to support absolute localization. Existing approaches rely on Monte Carlo Localization (i.e. Particle Filtering) \cite{thrun2001robust,winterhalter2015accurate, maffei2020global, ito2014w, ribacki2018vision, wang2019glfp, boniardi2019robot}, Stochastic Gradient Descent~\cite{li2020online}, and Voronoi segmentation~\cite{hobby2018method}. \cite{ribacki2018vision, maffei2020global} introduce \textit{Free-Space Density}, a measure of the free-space in a given circular radius around the current position used to support their Monte-Carlo Localization framework.
\cite{watanabe2020robust} use information from a depth camera in conjunction with generalized ICP. Further alternatives are given by \cite{maffei2020global}, \cite{winterhalter2015accurate}, and \cite{ito2014w} who use a depth sensor, an inertial RGB-D, or a combination of an RGB-D sensor and WiFi information, respectively. In contrast to our work, the above works primarily focus on the question of how to initialize the absolute pose, and map information is not given in the form of a floor plan prior.

An approach specifically tailored to tracking is given in \cite{el2005road}. However, the method again does not rely on floor plans as prior map information. The most similar works compared to ours are given by Boniardi et al. \cite{boniardi2017robust, boniardi2019pose}. They also utilize a CAD floor plan prior to support long-term navigation based on pose graph optimization or Bayes filtering. However, their work is different and outperformed by ours in three regards: First, their method addresses a three DoF estimation problem and uses only 2D LiDAR scans. This can easily lead to robustness issues, for example in the case of occlusions or out of plane motion. Second, they transform the floor plan into a discrete occupancy grid, which leads to a loss in accuracy. Third, their method is not as efficient as ours as demonstrated by a performance of only 12 Hz on a 4.00GHz i7-4790K CPU. In contrast, our method achieves up to 50Hz real-time performance on a 1.60GHz i5-8250U CPU.


\section{Correspondence Search in Floor Plans}
\label{sec:ANNF}

This section first reviews the pros and cons of different nearest neighbour search methods between point clouds and a floor plan-based map. After a clear problem definition, we introduce a novel approximate nearest neighbour field to efficiently retrieve point to floor plan element correspondences. We conclude the section with a validation test to prove its accuracy and efficiency. 

\subsection{Motivation}

Let $\mathcal{E}_{M}$ be a set that consists of parametric forms of geometric elements such as line segments, circular columns, and curve segments. It is obtained from a CAD floor plan through several pre-processing steps including reading of semiotic labels. Let $\mathcal{P}_L$ furthermore be a set of points captured during one LiDAR scan, and $\mathcal{P}_M$ be a set of 2D sampled points from $\mathcal{E}_{M}$. For convenience, here we assume that all points are 2D readings from a horizontal single-line LiDAR or sampled in the 2D floor-plan, i.e. $\textbf{p} \in \mathbb{R}^2$. The handling of 3D point clouds will be discussed later on. Suppose we have an estimated rotation matrix $\textbf{R}$ and translation vector $\textbf{t}$, a transformed point can be deduced as 
\begin{equation}
    \label{transformed point}
    \mathbf{p}_i^{\prime} = \textbf{R}\mathbf{p}_i + \textbf{t} \,, \ 
    \mathbf{p}_i \in \mathcal{P}_L \,.
\end{equation}
We then define the point-to-point registration as
\begin{equation}
    \label{point-to-point registration}
    \sigma^*_{\text{p2p}} = \min_{\textbf{R}, \textbf{t}} \sum_{\mathbf{p}_i \in \mathcal{P}_L}
                            || \pi(\mathbf{p}_i^{\prime}, \mathcal{P}_M), \mathbf{p}_i^{\prime} ||^2\,,
\end{equation}
where $\pi(\mathbf{p}_i^{\prime}, \mathcal{P}_M)$ finds the closest corresponding sample point from the set $\mathcal{P}_M$ given $\mathbf{p}_i^{\prime}$, and $||\cdot,\cdot||^2$ is defined as the squared Euclidean distance between two points.

The form of the above objective is similar to the classical ICP problem, which consists of registering the point sets of two LiDAR scans with adjacent timestamps \cite{besl1992method}. We converted a continuous representation into a discrete form by sampling points on line and curve segments in order to meet the required form for point-to-point registration. This is sub-optimal. Even if we may use a KD-tree to store and search the reference points in logarithmic time, the point-level data format inevitably incurs a loss in accuracy yet still can be significantly improved in terms of look-up time.

We omit the point sampling and define the point-to-element registration as
\begin{equation}
    \label{point-to-element registration}
    \sigma^*_{\text{p2e}} = \min_{\textbf{R}, \textbf{t}} \sum_{\mathbf{p}_i \in \mathcal{P}_L}
                            || \pi(\mathbf{p}_i^{\prime}, \mathcal{E}_M), \mathbf{p}_i^{\prime} ||^2 \,.
\end{equation}
$\pi(\mathbf{p}_i, \mathcal{E}_M)$ now finds potential correspondences by retrieving closest elements from the set $\mathcal{E}_M$ given the input point $\mathbf{p}_i$. Note that here and remainder of this paper, the definition of $||\cdot,\cdot||^2$ is more general and defined as the smallest distance between a point and any point sampled on another geometric element. The above distance measure has the advantage of behaving continuously as a function of the input sampling point, as it does not suffer from discretization noise. The challenge is now given by a data structure enabling efficient and robust retrieval of nearest points. In the following, we will introduce a novel Approximate Nearest Neighbour Field (ANNF) structure to support the function $\pi(\cdot)$ in Equation \ref{point-to-element registration} for efficient nearest neighbour matching.

\subsection{ANNF for geometric element retrieval}

ANNFs and their relation to more commonly used representations such as distance fields have been introduced in prior work such as \cite{zhou2018canny}. The theory about ANNFs contains two parts: construction and search. Other operations such as insertion, deletion, and traversal are not put into consideration in the present application, as we assume that the floor plan of a building is fixed.

To construct the ANNF, we first rasterize the input floor plan into multiple quadratic root fields with pre-set length. This makes sure that each field initially has the same length, and that---irrespectively of the size of the floor plan---a certain minimum resolution is guaranteed. Next, we iterate over every element, check which fields they are passing through, and note down the distance to the center of each field. For each field, the two elements with shortest distance to the field center are retained. In order to increase resolution, we then divide every field into four smaller quadratic subfields with only half the length of their parent field. The two nearest elements are again found using the above procedure. If a child subfield shares the same two nearest elements with its parent field, it will not be further divided and set as a leaf field. If all four subfields of the same parent field have the same two nearest neighbours as their parent's, the division operation will be withdrawn and the parent field will be set as a leaf field. Note that there are pathological cases in which this leads to approximations. However, these cases are rare and the above termination criteria are helpful in maintaining an overall small size for the tree. We furthermore store two nearest elements in each field, which helps to reduce occasional errors happening at the boundary of fields. We also define a maximum depth $d$ to control the overall size of the ANNF. Any subfield reaching this depth will no longer get divided.

Looking up in the ANNF is similar to a search through an adaptively-sampled quad-tree structure. Given a certain 2D point, we first locate the root field it lies in, which is a simple rounding operation. We then iteratively jump to child subfields until the corresponding leaf field is reached. The leaf field will output two nearest elements as potential nearest structure elements for the input point. Using the ANNF transfers computation time for calculating point-to-nearest-element distances to the field construction phase, which can be done offline in advance. The computational complexity of a look-up is reduced from $\mathcal{O}(|\mathcal{E}_M|)$ to $\mathcal{O}(d)$, where $|\mathcal{E}_M|$ represents the cardinality of set $\mathcal{E}_M$, and $d$ stands for the maximum depth of the ANNF, respectively.

\subsection{Performance Validation}
\begin{figure}[t]
  \centering
  \includegraphics[width=0.48\textwidth]{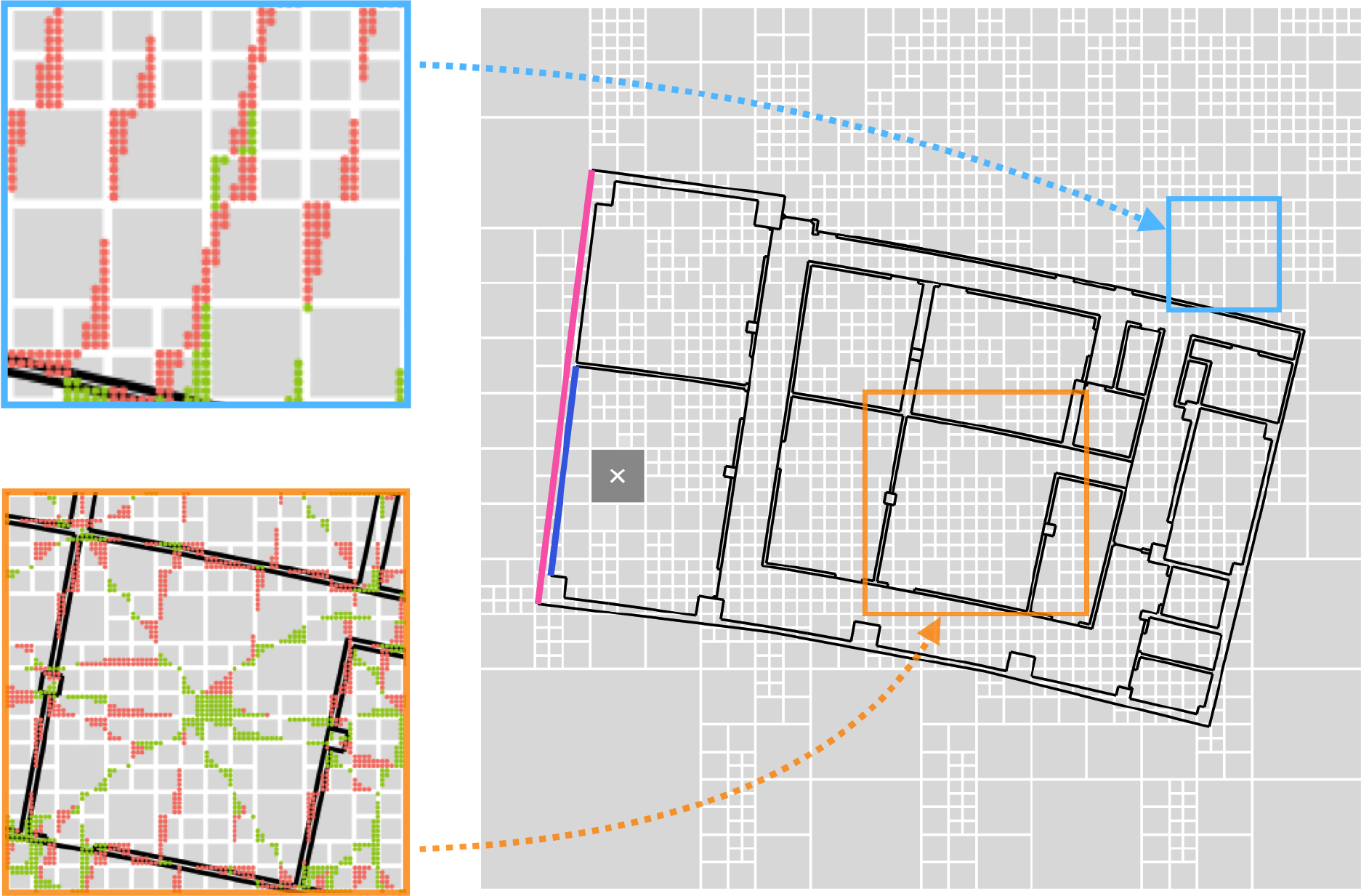}
  \caption{Approximate Nearest Neighbor Fields (ANNF) with at most $d = 5$ layers in depth. Every quadratic subfield stores two nearest elements from the floor plan. For example, the highlighted subfield marked with a cross stores a blue line segment as its nearest neighbor and a pink line segment as its second nearest neighbor. The data structure can be constructed up-front, and can be used as a look-up table with $\mathcal{O}(d)$ complexity. On the left, we show a performance validation as two zoom-in figures. The orange dot indicates a tested position that has the second nearest element from the ANNF as its closest element. The green dot indicates a location where the ANNF search did not include the ground truth closest element. An interactive demo can be found at \url{https://mpl-annf.github.io/}}
  \label{ANNF_diagram}
\end{figure}
The ANNF will only return an approximate result for correspondence matching. In order to analyze its accuracy, we make an exhaustive test on a floor plan denoted \textit{Corridor}. We construct a series of ANNFs with different depths. For each one, we uniformly sample about 40 million points within the boundaries of the environment, and record the retrieved first and second nearest elements from the ANNF. We also find the ground truth information about the nearest element by exhaustively calculating the distance to all elements in the floor plan for every sampled point. Quantitative results can be found in Table \ref{ANNF_Validation} and a visualization\footnote{For visualization purpose, the example in the figure only contains around 100 thousand points on a 5-layer ANNF.} is also provided in Fig. \ref{ANNF_diagram}, where orange dots indicate correct hits by the second nearest neighbor, and green dots are locations where none of the two approximate nearest neighbours corresponded to the ground truth element.

It is intuitively clear that accuracy rises as the maximum depth of the ANNF is increased. The search time for single points however remains in the same order, ranging from 0.16 to about 0.35ns. As can be observed from the visualization in Fig. \ref{ANNF_diagram}, most of the positions where the second nearest element is the best (orange dots) are lying on the angle bisector between two intersected elements. Furthermore, most of the samples where the nearest neighbour was entirely missed (green dots) are lying in a dense area where curves are approximated by multiple line segments, thus leading to an incorrect nearest neighbour but still an acceptable match for our registration. Note that a large portion of the samples where the nearest neighbour was missed are lying outside the boundaries of the floor plan. In practice, almost no samples will occur in this area, which is why it has very limited impact on the performance of our actual registration result.

In our experiments, we set a maximum depth of $d = 7$, which ensures a resolution of about 10cm, a reasonable choice given that the depth accuracy of common LiDAR sensors lies in the same order. This configuration enjoys high success rate and low time consumption for the nearest neighbour search.

\begin{table}[t]
  \caption{Performance Validation for ANNF}
  \label{ANNF_Validation}
  \begin{center}
  \begin{tabular}{ccccc}
    \toprule
    \multirow{2}{*}{Depth} & Leaf field & Hit by   & Hit by 1st  & Search   \\
                           & Length[cm] & 1st elmt & or 2nd elmt & Time[ns] \\
    \midrule
    3 & 150.00 & 61.61\% & 76.10\% & 0.24 \\
    4 & 75.00 & 76.08\% & 87.85\% & 0.16 \\
    5 & 37.50 & 86.29\% & 93.61\% & 0.24 \\
    6 & 18.75 & 91.96\% & 96.24\% & 0.18 \\
    7 &  9.38 & 94.90\% & 97.28\% & 0.25 \\
    8 &  4.69 & 96.46\% & 98.84\% & 0.35 \\
    \bottomrule
  \end{tabular}
  \end{center}
\end{table}


\section{Floor Plan-assisted Indoor Localization}

\begin{figure}[t]
  \centering
  \includegraphics[width=0.47\textwidth]{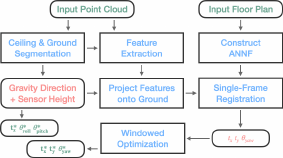}
  \caption{Block diagram illustrating the full pipeline of the proposed Floor Plan-assisted Indoor Localization (FP-Loc). The system uses readings from a 3D LiDAR and prior knowledge of a floor plan. It generates a six DoF trajectory from a two-stage optimization procedure.}
  \label{pipeline}
\end{figure}

An overview of the whole pipeline along with inputs, outputs, intermediate results, and six sub-modules is presented in Fig. \ref{pipeline}. The ANNF is constructed upfront from an input floor plan (cf. Section \ref{sec:ANNF}). At online stage, the system will first segment the point clouds into ceiling, ground, and wall points by a robust plane fitting algorithm (cf. Section \ref{Segmentation}). The ceiling and ground plane parameters implicitly reveal the gravity direction and the height of the sensor, which determines the first three Degrees of Freedom (DoF) of the trajectory $[t_\text{z}, \theta_{\text{roll}}, \theta_{\text{pitch}}]$. Next, the wall points are used to extract features which are furthermore projected onto the horizontal plane using the gravity direction (cf. Section \ref{Feature Extraction and Projection}). The projected features are finally used to look-up nearest geometric elements in the ANNF and conduct a two-stage optimization of the pose. The first stage aligns a single frame and judges whether or not it constitutes a keyframe. The second stage then jointly aligns multiple frames taking into account pair-wise regularization terms. This pose-graph optimization produces the remaining three DoF of the trajectory $[t_\text{x}, t_\text{y}, \theta_{\text{yaw}}]$ (cf. Section \ref{Optimization}).

\subsection{Ceiling and Ground Plane Segmentation}
\label{Segmentation}

We denote the set of points scanned during a full turn of one of the LiDAR's rotating laser beams as a ring.
Before we introduce the core idea behind the segmentation of the ceiling and ground planes of an indoor environment, we first make two important statements about the nature of LiDAR scans:

\begin{itemize}
\item If the LiDAR is mounted horizontally on a platform, the top ring of each LiDAR scan will have the largest possibility of hitting the ceiling owing to the fact that its rays have the largest elevation angle compared to the rays of all other rings.
\item If the LiDAR is mounted horizontally on a platform, on one and the same ring, points with larger distance along the ray have a higher possibility of belonging to the ceiling. On the other hand, if a scanning ray is obstructed, the distance along the ray and thus also the distance to the sensed point will be smaller.
\end{itemize}

We start by using these insights to design our ceiling segmentation algorithm. Initial plane parameters are obtained by taking the points with the largest depth-along-ray for the top $K_r$ rings. We then obtain an initial guess of the ceiling plane parameters by simple linear regression. We finally proceed to an M-estimator that robustly fits the ceiling plane by using all points on the upper half rings. The resulting closed-form plane parameters easily reveal the normal vector of this plane which is equivalent to the direction of gravity. We may furthermore extract the Euclidean distance from the origin to the plane, and use it as information to calculate any height variations of the sensor. Note that the strategy works well in any environment in which the dominant part of the ceiling is made up by a horizontal plane, which includes a majority of indoor spaces. In summary, the first three DoF of the trajectory---$[t_\text{z}, \theta_{\text{roll}}, \theta_{\text{pitch}}]$---are retrieved by a plane fitting strategy.

The previous statements and the resulting algorithm can be easily modified towards ground plane segmentation by vertical flipping (i.e. we start from the bottom ring instead of the top ring). However, the result of ground plane segmentation is only used to further support the singling out of points on vertical walls and pillars, not to refine the orientation or vertical position of the sensor. This strategy is supported by our real world setup, in which the LiDAR is mounted in a relatively high position on a large tripod. Owing to this positioning, the sensor simply samples more points on the ceiling and fewer points on the ground. Thus, using the ceiling points to estimate the gravity direction and the vertical position is a more robust solution.

\subsection{Feature Extraction and Projection}
\label{Feature Extraction and Projection}

We use feature extraction techniques adopted by LOAM \cite{zhang2014loam} in order to turn the segmented wall points into robust and distinctive points. Our algorithm classifies 3D points into two feature categories, which are corner points and surface points. They are distinguished by evaluating local curvature or smoothness in individual scan rings. Both corner features and surfaces features are later on projected onto the horizontal ground plane using the gravity direction. Note that point clouds are easily compensated for roll and pitch angles by using the gravity direction.

Although projected surface points and corner points are weighted equally during the later registration, the projected corner points can be used as anchors to further classify the projected surface points into subgroups. More specifically, if all the projected surface points between two corner points constitute a line segment (judged by principal component analysis), the corresponding points are labelled as belonging to the same geometric element. This information is used during optimization to enforce consistent matching of those points to one and the same geometric element from the map. If the projected surface points do not satisfy this constraint, they are labelled as \textit{unclassified} and remain able to find independent correspondences.

\subsection{ANNF-based Pose Graph Optimization}
\label{Optimization}

\begin{figure}[t]
  \centering
  \includegraphics[width=0.35\textwidth]{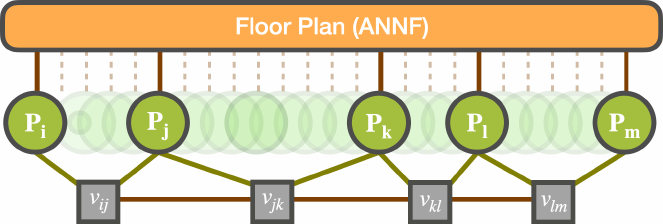}
  \caption{Factor graph illustrating the optimization problems during localization. The green nodes show the poses corresponding to each LiDAR scan, the orange rectangle is the prior knowledge given by the floor plan (represented in the form of an ANNF), and the grey squares are approximating linear/angular velocities based on the relative pose and the elapsed time between key poses. The latter are regularized over time to smoothify the motion. The first stage of our optimization---\textit{single-frame registration}---optimizes only a single pose based on the floor plan prior. We then consult the distance between the optimized pose and the previous key pose to perform keyframe selection. If it is not a key pose, the current scan will be discarded. If it is a key pose, the new frame is added and we proceed to the second stage of our optimization, which is \textit{windowed optimization} over multiple keyframes. In addition to the floor plan prior, this pose-graph optimization also considers the regularization terms on linear/angular velocities.}
  \label{factor_graph}
\end{figure}

Our localization is based on pose graph optimization and contains two steps: \textit{Single-Frame Registration} and \textit{Windowed Optimization} for pose refinement. The factor graph is presented and explained in Fig. \ref{factor_graph}.

For \textit{single-frame registration}, we first define a transformed point,
\begin{equation}
    \mathbf{p}_n^{\prime} = \textbf{R}(\theta_{\text{yaw}})\mathbf{p}_n + \textbf{t}(t_\text{x}, t_\text{y}) \,, \ \mathbf{p}_n \in \mathcal{F} \,,
\end{equation}
where the estimated three parameters $t_\text{x}$, $t_\text{y}$ and $\theta_{\text{yaw}}$ form the translation vector \textbf{t} and the rotation matrix $\textbf{R}$, respectively. $\mathcal{F}$ refers to the set of all projected 2D feature points. The objective is given by
\begin{equation}
    \sigma^*_{\text{sfr}}(i) = 
    \min_{t_\text{x}, t_\text{y}, \theta_{\text{yaw}}} \sum_{\mathbf{p}_n \in \mathcal{F}}
    || \textbf{ANNF}(\mathbf{p}_n^{\prime}, \mathcal{E}_M), \mathbf{p}_n^{\prime} ||^2 \,,
\end{equation}
where $\textbf{ANNF}(\cdot)$ is our nearest neighbour field-based search for geometric elements corresponding to a given 2D point. Here, $i$ indicates the $i$-th frame of the LiDAR scan.

Next, the optimized parameters $t_\text{x}, t_\text{y}$ along with $t_\text{z}$ are subjected to a distance check with respect to the previous key pose. If the distance threshold is not exceeded, the scan is dropped after single-frame registration. If the distance reaches a certain threshold, the scan is selected as a key frame, and we proceed with \textit{windowed optimization}. We first define the linear velocity and angular velocity between two consecutive key poses as  
\begin{equation}
    \small
    v_\text{x}^j = \frac{t_\text{x}^{j+1} - t_\text{x}^{j}}{\tau_{j+1} - \tau_{j}} \,, \ 
    v_\text{y}^j = \frac{t_\text{y}^{j+1} - t_\text{y}^{j}}{\tau_{j+1} - \tau_{j}} \text{ and } 
    \omega^j = \frac{\theta_\text{yaw}^{j+1} - \theta_\text{yaw}^{j}}{\tau_{j+1} - \tau_{j}} \,,
\end{equation}
where $\tau_j$ refers to the time of the $j$-th keyframe. Note that the $j$-th keyframe is defined as the first keyframe in a window of $W$ most recent keyframes. Thus, the objective is defined as
\begin{equation}
    \sigma^*_{\text{wo}}(j, \cdots, j+W) = \min_{t_\text{x}^j, t_\text{y}^j, \theta^j_{\text{yaw}}, \cdots} 
                                           \sum_{m = j}^{j+W} \sigma^*_{\text{sfr}}(m) + \nonumber
\end{equation}
\begin{equation}
    \small
    \sum_{m = j}^{j+W-1} \alpha ||v_\text{x}^{m+1} - v_\text{x}^{m}||^2 + \alpha ||v_\text{y}^{m+1} - v_\text{y}^{m}||^2 + \beta ||\omega^{m+1} - \omega^{m}||^2,
\end{equation}
which includes $W$ distinct \textit{single-frame registration} objectives but additionally regularizes first-order approximations of linear and angular velocities over time, thus smoothing the overall trajectory result. $\alpha$ and $\beta$ are hyper-parameters. Each key pose is optimized $W$ times, and the final position and orientation $[t_\text{x}, t_\text{y}, \theta_{\text{yaw}}]$ in the results is the one after the final optimization, just before it leaves the optimization window. In our experiments, we set $\alpha = 1.1$, $\beta = 0.9$, and $W = 10$.


\section{Experimental Evaluation}

We perform two sets of experiments to validate our algorithm's accuracy and efficiency, and to qualitatively and quantitatively compare our method against a popular SLAM alternative (LeGO-LOAM \cite{shan2018lego}). Note that our aim here is not to introduce an alternative to SLAM, but merely to demonstrate how FP-Loc is effective in preventing the accumulation of drift.

\subsection{Datasets and implementation}

We use our own datasets captured in unfurnished rooms or low-texture maze-like long corridors. Data is captured by a tripod placed on a dolly and carrying an Ouster OS-1 64-line LiDAR (running at 10Hz), an XSense Mti-30 9-axis IMU (running at 200Hz), and a Grasshopper RGB camera (running at 30Hz, used for visualization purposes). The entire sensor set has been calibrated and synchronized in advance. All our methods are implemented in C++ and executed using the robot operating system (ROS) under Ubuntu. All experiments are conducted on a laptop with an Intel Core i5-8250U 1.6 GHz CPU and 8GB RAM. Ground truth for each data sequence is generated by running LIO-SAM with loop closure activated \cite{shan2020lio}.

\subsection{Results in unfurnished rooms}

Our first experiments are conducted in a $200m^2$ room with two doors and windows located on opposing walls. The room furthermore contains square and circular pillars. The room is recently constructed and therefore has limited furnishing and almost no visual texture.

\begin{figure}
  \centering
  \includegraphics[width=0.38\textwidth]{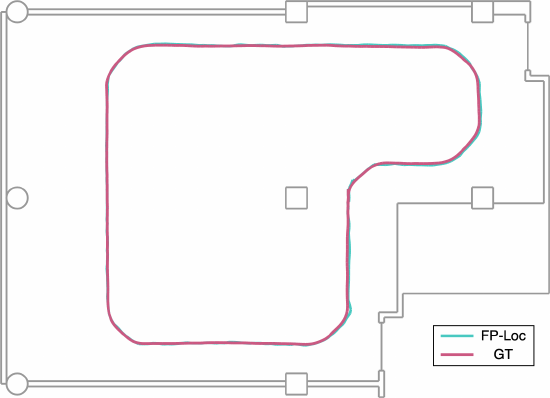}
  \caption{A bird's eye view onto a trajectories captured in an unfurnished room (large loop with fast motion).}
  \label{Unfurnished_RM_Big_Loop_Fast}
  \end{figure}

We have performed tests over nine sequences under different conditions: \textit{Small Loop}, \textit{Large Loop}, and \textit{No Loop}, with different speed. Mean Relative Pose Errors (RPE) and Mean Absolute Trajectory Errors (ATE) \cite{sturm2012benchmark} before and after \textit{windowed optimization} are indicated in Table \ref{Unfurnished_RM_Table}. Bird's eye views of the optimized trajectories with comparison to ground truth are indicated in Fig. \ref{Unfurnished_RM_Big_Loop_Fast}.

As can be observed, we obtain drift-free results very close to ground truth, and our pose-graph optimization module helps to smooth the trajectory and improve RPE and ATE scores. Averaging runtime ranges between 35 and 50Hz, which is up to 5x faster than real time given the LiDAR sampling frequency of 10Hz.

\subsection{Long corridor}
\begin{table*}[t]
  \caption{Accuracy Results on the unfurnished room sequence}
  \label{Unfurnished_RM_Table}
  \begin{center}
  \begin{tabular}{ccccccccc}
    \toprule
    \multirow{2}{*}{Data} & \multirow{2}{*}{Motion Type} & Max. Speed & Avg. Speed & Length
    & \multicolumn{2}{c}{\underline{\smash{Single Frame Registration}}} & \multicolumn{2}{c}{\underline{\smash{Windowed Optimization}}} \\
    & & [m/s] & [m/s] & [m] & RPE[cm] & ATE[cm] & RPE[cm] & ATE[cm] \\
    \midrule
    \multirow{3}{*}{Small Loop} & Fast   & 0.8719 & 0.6236 & 25.797 & 2.7338 & 4.8188 & \textbf{2.6900} & \textbf{4.6736} \\
                                & Medium & 0.5641 & 0.4090 & 25.849 & 3.3473 & 4.9413 & \textbf{3.3470} & \textbf{4.6676} \\
                                & Slow   & 0.4425 & 0.3109 & 25.714 & \textbf{3.3062} & 5.3061 & 3.3132 & \textbf{5.2476} \\
    \midrule
    \multirow{3}{*}{Large Loop} & Fast   & 0.9213 & 0.7253 & 46.725 & 3.1447 & 5.9842 & \textbf{3.1234} & \textbf{5.8678} \\
                                & Medium & 0.6878 & 0.5357 & 46.773 & 3.0335 & 5.3048 & \textbf{3.0256} & \textbf{5.1823} \\
                                & Slow   & 0.4742 & 0.3552 & 46.421 & 3.7812 & 6.0282 & \textbf{3.7555} & \textbf{5.9972} \\
    \midrule
    \multirow{3}{*}{No Loop}    & Fast   & 0.8458 & 0.6926 & 25.287 & 1.7505 & 5.9954 & \textbf{1.7013} & \textbf{5.5408} \\
                                & Medium & 0.5277 & 0.3904 & 25.322 & 2.1412 & 4.4469 & \textbf{2.1360} & \textbf{4.3437} \\
                                & Slow   & 0.4028 & 0.2954 & 25.208 & 2.0445 & 4.9362 & \textbf{2.0422} & \textbf{4.9270} \\
    \bottomrule
  \end{tabular}
  \end{center}
\end{table*}
\begin{table*}[t]
  \caption{Performance Comparison on the long corridor sequence}
  \label{Corridor_Table}
  \begin{center}
  \begin{tabular}{ccccccccc}
    \toprule
    \multirow{2}{*}{Data} & \multirow{2}{*}{Motion Type} & Max. Speed & Avg. Speed & Length 
    & \multicolumn{2}{c}{\underline{\smash{\ \ \ \ LeGO-LOAM\ \ \ \ \ }}} & \multicolumn{2}{c}{\underline{\smash{\ \ \ \ \ \ \ \ FP-Loc\ \ \ \ \ \ \ \ \ }}} \\
    & & [m/s] & [m/s] & [m] & RPE[cm] & ATE[cm] & RPE[cm] & ATE[cm] \\
    \midrule
    \multirow{3}{*}{Loop} & Fast   & 1.2678 & 1.0487 & 81.448 & \textbf{2.6785} & 132.6895 & 3.3826 & \textbf{24.3258} \\
                          & Medium & 0.9296 & 0.7777 & 81.789 & 9.2715 & 131.7793 & \textbf{3.1069} & \textbf{23.8171} \\
                          & Slow   & 0.6603 & 0.5547 & 82.072 & 7.5712 & 129.9051 & \textbf{3.0835} & \textbf{24.8676} \\
    \bottomrule
  \end{tabular}
  \end{center}
\end{table*}
\begin{figure}
  \centering
  \includegraphics[width=0.42\textwidth]{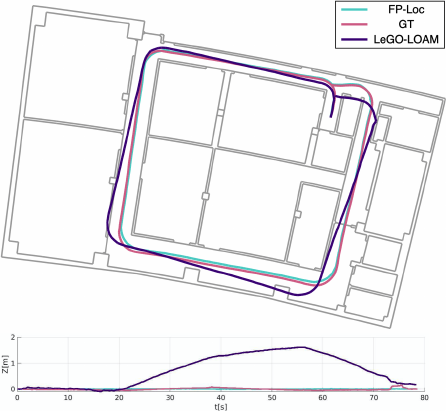}
  \caption{Top: A bird's eye view onto trajectories estimated with fast motion in a corridor environment. Bottom: Details along Z-axis.}
  \label{Corridor_Loop_Fast}
\end{figure}

In a final experiment, we aim at demonstrating the drift-free localization ability of our algorithm under challenging conditions. The dataset is recorded in a low-texture underground environment with four highly self-similar bended corridors. The width of each corridor lies between 1.5m and 3m, and their length is roughly 20m each. We again perform tests for three different motion speeds, \textit{Slow}, \textit{Medium}, and \textit{Fast}. In order to prove the advantage of our absolute registration technique, we have also tested LeGO-LOAM with loop closure deactivated (the latter is necessary to prevent unwanted loop closures due to the highly self-similar environment). Table \ref{Corridor_Table} indicates the corresponding RPE and ATE compared against LIO-SAM~\cite{shan2020lio} based ground truth. We also show a bird's eye view of both our and ground truth trajectory in Fig. \ref{Corridor_Loop_Fast}. The averaging frame rate for this dataset lies between 15 and 20Hz, which still exceeds real-time.


\section{Conclusions}

We presented a novel approach for continuous, floor plan based, full six degree-of-freedom localization of 3D LiDARs. The introduction of an efficient nearest-neighbour field for geometric floor plan elements enables the construction of smooth, continuously differentiable residual terms, and thus outperforms more conventional discrete distance fields. Our chosen data structure furthermore enables highly efficient execution that potentially runs on computationally constrained hardware. We believe that the introduced algorithm represents a highly interesting alternative to existing indoor localization schemes, which all depend on the installation of dedicated hardware. Our future efforts consist of increasing the generality of the method by adding more complex, inertial-assisted ceiling segmentation and artificial intelligence modules aiming at the detection and active removal of objects and clutter in the measured point clouds.

\section*{Acknowledgement}

The authors would like to thank the funding sponsored by the Natural Science Foundation of Shanghai (grant number: 22ZR1441300), as well as the support provided by our industry partner Stereye Intelligent Technology.


\addtolength{\textheight}{-8.4cm} 



\end{document}